\documentclass[11pt,a4paper]{article}
\usepackage[hyperref]{acl2017}
\usepackage{times}
\usepackage{latexsym}
\usepackage{amssymb}
\usepackage{graphicx}
\usepackage{stmaryrd}

\usepackage{tikz} 			 
\usepackage{pgfplots}		 

\pgfplotsset{compat=1.10}	 
\usepgfplotslibrary{colorbrewer} 

\usepackage[]{algorithm2e}	 

\usepackage{caption}		 

\usepackage{amsmath}
\usepackage{natbib}
\usepackage[cal=boondox,scr=rsfs]{mathalfa}
\usepackage{textgreek}\usepackage{url}
\usepackage{amssymb,amsmath}

\newcommand{\comment}[1]{}

\title{The Complex Negotiation Dialogue Game}

\author{Romain Laroche \\
  Microsoft Maluuba, Montr\'eal, Canada \\
  {\tt romain.laroche@microsoft.com} \\}

\date{}

\begin{document}
\maketitle

\begin{abstract}
This position paper formalises an abstract model for complex negotiation dialogue. This model is to be used for the benchmark of optimisation algorithms ranging from Reinforcement Learning to Stochastic Games, through Transfer Learning, One-Shot Learning or others. 
\end{abstract}

\section{Introduction}
A negotiation is defined as a \textit{bargaining process between two or more parties (each with its own aims, needs, and viewpoints) seeking to discover a common ground and reach an agreement to settle a matter of mutual concern or resolve a conflict}. From a dialogue point of view, one distinguishes negotiation dialogue from standard dialogue by the mutual sharing of information\footnote{whereas standard dialogue mainly relies on discovering the user information or intent,}, by its required user adaptation\footnote{whereas standard dialogue, such as form filling applications, is rather indifferent to the user's characteristics,}, and by the non-stationarity induced by its non fully cooperative structure: the user and system objectives correlate but also differ to some extent, and they are consequently adversely co-adapting.

Research on negotiation dialogue experiences a growth of interest. At first, Reinforcement Learning~\cite{Sutton1998}, the most popular framework for dialogue management in dialogue systems~\cite{Levin1997,Laroche2009,Lemon2012}, was applied to negotiation with mitigated results~\cite{English2005,Georgila2011,Lewis2017}, because the non-stationary policy of the opposing player prevents those algorithms from converging consistently. Then, Multi-Agent Reinforcement Learning~\cite{Bowling2002} was applied but also with convergence difficulties~\cite{Georgila2014}. Finally, recently, Stochastic Games~\cite{Shapley1953} were applied successfully~\cite{Barlier2015}, with convergence guarantees, but only for zero-sum games, which is inconsistent with dialogue since most tasks are cooperative.

Here, we extend \cite{Laroche2017b}'s abstraction of the negotiation dialogue literature applications: \cite{diEugenio2000,English2005} consider sets of furniture, \cite{Afantenos2012,Efstathiou2014,Georgila2014,Litman2016,Lewis2017} resource trading, and \cite{Putois2010,Laroche2011,ElAsri2014,Genevay2016,Laroche2017a} appointment scheduling. Indeed, these negotiation dialogue problems are cast into a generic agreement problem over a shared set of options. The goal for the players is to reach an agreement and select an option. This negotiation dialogue game can be parametrised to make it zero-sum, purely cooperative, or general sum. However, \cite{Laroche2017b} only consider elementary options: they are described through a single entity. 

We formalise in this paper the game for options that are compounded in the sense that they are characterised by several features. For instance, \textit{Tuesday morning} is defined by two features: the day and the moment of the day. Considering compounded options naturally leads to richer expressions, and therefore to a larger set of actions: \textit{I'm available whenever on Tuesday}, or \textit{I'd prefer in the afternoon}. Since the options are uttered in a compounded way, as opposed to their elementary definition in \cite{Laroche2017b}, the state representation also becomes more complex. This extension allows more realistic dialogues, and more challenging Reinforcement Learning, Multi-Agent Reinforcement Learning, and Stochastic Games policy training.

\section{The Negotiation Dialogue Game}
\label{sec:negotiation}
This section recalls the negotiation dialogue game as described in \cite{Laroche2017b}. The goal for each participant is to reach an agreement. The game involves a set of $m$ players $\mathcal{P} = \{\mathcal{p}^i\}_{i\in\left[1,m\right]}$. With $m>2$, the dialogue game is said to be multi-party~\cite{Asher2016,Litman2017}. The players consider $n$ options (in resource trading, an option is an exchange proposal, in appointment scheduling, it is a time-slot), and the cost to agree on an option $\tau$ is $c_{\!\tau}^i$ randomly sampled from distribution $\delta^i \in \Delta_{\mathbb{R}^+}$ to agree on it. Players also have a utility $\omega^i \in \mathbb{R}^+$ for reaching an agreement. For each player, a parameter of cooperation with the other players $\alpha^i \in \mathbb{R}$ is introduced. As a result, player $\mathcal{p}^i$'s immediate reward at the end of the dialogue is:
\begin{equation}
R^i(s_T^i) = \omega^i-c_{\!\tau}^i + \alpha^i \sum_{j\neq i} (\omega^{j}-c_{\!\tau}^j)
\end{equation}
where $s_T^i$ is the last state reached by player $\mathcal{p}^i$ at the end of the dialogue, and $\tau$ is the agreed option. If players fail to agree, the final immediate rewards $R^i(s_T^i)=0$ for all players $\mathcal{p}^i$. If at least one player $\mathcal{p}^j$ misunderstands and agrees on a wrong option $\tau^j$ which was not the one proposed by the other players, this is even worse: each player $\mathcal{p}^i$ gets the cost of selecting option $\tau^i$ without the reward of successfully reaching an agreement:
\begin{equation}
R^i(s_T^i) = -c_{\!\tau^i}^i - \alpha^i \sum_{j\neq i} c_{\!\tau^j}^j \label{eq:badSchedule}
\end{equation}

The values of $\alpha^i$ give a description of the nature of the players, and therefore of the game as modelled in game theory \cite{Shapley1953}. If $\alpha^i<0$, player $\mathcal{p}^i$ is said to be antagonist: he has an interest in making the other players lose. In particular, if $m=2$ and $\alpha^1=\alpha^2=-1$, it is a zero-sum game.  If $\alpha^i=0$, player $\mathcal{p}^i$ is said to be self-centred: he does not care if the other player is winning or losing. Finally, if $\alpha^i>0$, player $\mathcal{p}^i$ is said to be cooperative, and in particular, if $\forall i \in\left[1,m\right]$, $\alpha^i=1$, the game is said to be fully cooperative because $\forall (i,j) \in\left[1,m\right]^2$, $R^i(s_T^i) = R^j(s_T^j)$.

From now on, and until the end of the article, we suppose that there are only $m=2$ players: a system $\mathcal{p}_s$ and a user $\mathcal{p}_u$. They act each one in turn, starting randomly by one or the other. They have four possible actions. \textsc{Accept}$(\tau)$ means that the user accepts the option $\tau$ (independently from the fact that $\tau$ has actually been proposed by the other player; if it has not, this induces the use of Equation \ref{eq:badSchedule} to determine the reward). This act ends the dialogue. \textsc{RefProp}$(\tau)$ means that the user refuses the proposed option and proposes instead option $\tau$. \textsc{AskRepeat} means that the player asks the other player to repeat his proposition. And finally, \textsc{EndDial} denotes the fact that the player does not want to negotiate anymore, and terminates the dialogue.

Understanding through speech recognition of system $\mathcal{p}_s$ is assumed to be noisy with a sentence error rate $SER_s^u$ after listening to a user $\mathcal{p}_u$: with probability $SER_s^u$, an error is made, and the system understands a random option instead of the one that was actually pronounced. In order to reflect human-machine dialogue reality, a simulated user always understands what the system says: $SER_u^s=0$. We adopt the way \cite{Khouzaimi2015} generates speech recognition confidence scores: $score_{reco} = \frac{1}{1+e^{-X}} \text{ where } X \sim \mathcal{N}(c,0.2)$ given a user $\mathcal{p}_u$, two parameters $(c_{\!\bot}^u,c_{\!\top}^u)$ with $c_{\!\bot}^u<c_{\!\top}^u$ are defined such that if the player understood the right option, $c=c_{\!\top}^u$ otherwise $c=c_{\!\bot}^u$. The further apart the normal distribution centres are, the easier it will be for the system to know if it understood the right option, given the score. 

\section{Allowing compounded options}
This section extends the negotiation dialogue game recalled in Section \ref{sec:negotiation} with compounded options. Each option $\tau$ is now characterised by a set of $\ell$ features: $\tau = \{f^k_\tau\}_{k\in\llbracket 1,\ell\rrbracket}$, with $f^k_\tau\in\mathcal{F}^k$. Not all feature combinations might form a valid option, but for the sake of simplicity, we consider that the set of the $n$ options contain all of them and that the cost for inconsistent ones is infinite. This way, we can express that an option is invalid but that the user is not aware of it.

The cost of an option needs to be revisited consequently. The costs of two options that only differ by a feature are similar in general. Without loss of generality, we define the cost of one player $\mathcal{p}^i$ for agreeing on a given option $\tau$ as follows:
\begin{equation}
c^i_\tau = \hat{c}^i_\tau + \sum_{k=1}^\ell \hat{c}^i_k,
\end{equation}
where $\hat{c}^i_k$ is the cost of agreeing on feature $f^k$ and $\hat{c}^i_\tau$ is the cost for selecting this option in particular. In an appointment scheduling negotiation task, the feature related costs $\hat{c}^i_k$ can generally be considered as null: there is no correlation between being booked on Monday morning and being available on Tuesday morning. Most of the constraints are therefore expressed in the $\hat{c}^i_\tau$ term. On the opposite, in a furniture set application, the preferences are expressed on specific features of the furniture: colour, price, etc. In this case, the constraints mainly lie in $\hat{c}^i_k$ terms. 

The option-as-features definition naturally induces new ways of expressing one's preferences over the option set. The \textsc{PropFeatures}($f^{k_1}, f^{k_2}, \dots$) dialogue act replaces the previously defined \textsc{RefProp}($\tau$): it means that the speaker wants the $k_1^{\text{th}}, k_2^{\text{th}}, \dots$ features to be set to values $f^{k_1}, f^{k_2}, \dots$ \textsc{AskRepeat} still asks to repeat the whole last utterance, but its partial version is added: \textsc{AskPartialRepeat}($k_1, k_2, \dots$) consists in asking to repeat values of features {$k_1, k_2, \dots$} \textsc{Accept} still accepts the last grounded option, but it can only be performed once all features have been grounded. Its partial version is also introduced: \textsc{PartialAccept}($k_1, k_2, \dots$) determines an agreement on the last grounded value of features $k_1, k_2, \dots$

The compounded options imply complex actions, which in turn imply a complex understanding model: the sentence level understanding rate and score need to be extended. The sentence error rate $SER_s^u$ is therefore replaced with a feature error rate $FER_s^u$. The same speech recognition confidence score generation is used at the feature level, meaning that, at each \textsc{PropFeatures}, \textsc{AskPartialRepeat}, and \textsc{PartialAccept} acts, the player receives an array of feature values (or feature names), each associated with a confidence score.

\section{Potential use of the complex negotiation dialogue game}
\cite{Genevay2016} already used the simple negotiation dialogue game to study Knowledge Transfer for Reinforcement Learning \cite{Taylor2009,Lazaric2012} applied to dialogue systems \cite{Gasic2013,Casanueva2015}. It appears in this paper that the optimal policies are rather simple. Making the interaction process more intricate and more reality reflecting allows to put the computational tractability of the methods to the test. Following the same purpose, one-shot learning~\cite{Schaal1997,FeiFei2006} may also be used for negotiation dialogues.

Cooperative co-adaptation in dialogue has been tackled only in one previous article: \cite{Chandramohan2012}. Similarly, but for the adversary case, the negotiation dialogue game offers a good empirical test bed for a generalisation to the general-sum games of \cite{Barlier2015}. 

We believe that this line of research is complementary with the more applied one of \cite{Lewis2017} that work on real human dialogues and are more focused on dealing with natural language within a negotiation task. Their mitigated results indicate that negotiation generalisation over simulated users to real users is difficult, even when the simulated user is trained on human data.



\bibliographystyle{acl}
\bibliography{biblio}

\end{document}